\documentclass[11pt]{article}

\usepackage[final]{acl}

\usepackage{times}
\usepackage{latexsym}

\usepackage[T1]{fontenc}

\usepackage[utf8]{inputenc}

\usepackage{microtype}

\usepackage{inconsolata}

\usepackage{graphicx}

\usepackage{amsmath,amssymb}
\usepackage{booktabs}   
\usepackage{multirow}   
\usepackage{arydshln}   
\usepackage[table]{xcolor}
\usepackage{adjustbox}
\usepackage{graphicx}
\usepackage{subcaption}
\usepackage{bm}

\newcommand{\blanksymbolfootnote}[1]{%
  \renewcommand{\thefootnote}{}
  \footnote{#1}%
  \setcounter{footnote}{0} 
  \renewcommand{\thefootnote}{\arabic{footnote}}
}

\title{CAT: Confidence-Adaptive Thinking for Efficient Reasoning of Large Reasoning Models}

\author{
  Qizhi Jiang$^{1}$ \quad Shuo Wang$^{1}$ \quad Pei Ke$^{1,2,*}$ \quad Yuhang Song$^{1}$ \quad Ke Qin$^{1,2}$\\
  $^{1}$Laboratory of Intelligent Collaborative Computing, \\ University of Electronic Science and Technology of China, Chengdu, China \\
  $^{2}$Ubiquitous Intelligence and Trusted Services Key Laboratory of Sichuan Province \\
  \texttt{\{jiangqizhi, 202422900227\}@std.uestc.edu.cn, kepei@uestc.edu.cn} \\
  \texttt{songyuhang@std.uestc.edu.cn, qinke@uestc.edu.cn}
}

\begin{document}
\maketitle

\blanksymbolfootnote{$^*$Corresponding author.}

\begin{abstract}
Large Reasoning Models (LRMs) have achieved remarkable success on complex tasks by leveraging long chain-of-thought (CoT) trajectories, yet they frequently exhibit overthinking on simple queries, resulting in significant token overhead and reduced inference efficiency. However, existing compression methods predominantly apply uniform length reduction or rely on coarse-grained difficulty estimation, often leading to performance degradation on difficult problems. To address this limitation, we propose Confidence-Adaptive Thinking (CAT), a framework that incorporates the model's intrinsic self-certainty signals as confidence into the preference optimization process, which autonomously modulates reasoning lengths based on problem difficulty. Experimental results show that CAT consistently outperforms state-of-the-art baselines on reasoning accuracy across multiple benchmarks on different base models. Our work enables LRMs to effectively compress confident responses while deliberating on uncertain ones, offering a potentially robust solution for balancing accuracy and latency in practical industrial scenarios.

\end{abstract}

\section{Introduction}

Recently, large reasoning models (LRMs) have rapidly emerged and made substantial progress on complex natural language processing (NLP) tasks,
as exemplified by OpenAI-o1 \cite{Openai-o1} and DeepSeek-R1 \cite{Deepseek-r1}. 
These models are equipped with the ability to generate long reasoning chains, 
demonstrating strong potential on challenging reasoning problems such as mathematical competitions \cite{xu2025survey}. 
However, while LRMs heavily rely on long chain-of-thought (CoT) traces to perform well on difficult tasks, they tend to produce redundant reasoning and self-reflection for simple inputs, incurring pronounced overthinking and token overhead \cite{overthink,Feng_survey,Liu_survey,Sui_survey}. This behavior leads to verbose thought chains that increase computation cost and reduce overall inference efficiency.
Accordingly, how to enable LRMs to dynamically adjust token consumption based on the input difficulty has attracted increasing attention,
determining the practical industrial usability of LRMs in terms of the balance between accuracy and latency \cite{DAST}.

Most of the existing approaches focus on
reasoning compression and length control predominantly, which treat shortening reasoning chains as the primary objective \cite{Qu_survey} and apply a uniform reduction of reasoning tokens to all the queries \cite{xia-etal-2025-tokenskip,overthink,cot-valve,Munkhbat}. While such methods can substantially decrease generation length, they often incur non-trivial performance degradation on difficult problems, since complex tasks still require sufficient reasoning depths and lengths to sustain accurate answers \cite{s1_simple,Zeng_24}.
Another line of work resorts to difficulty-adaptive reasoning to mitigates the imbalance between overthinking for easier instances and underthinking for harder ones. This category of methods tends to dynamically adjust the budget of output tokens based on the model performance \cite{DAST}.

However, existing works on adaptive reasoning 
still face a severe challenge of coarse-grained difficulty estimation. Current methods utilize the accuracy of model outputs to measure the problem difficulty and roughly determine the output length \cite{DAST}. We argue that this coarse-grained estimation heavily relies on external labels and provides a partial assessment merely on the answer, rather than measuring the quality of the whole reasoning chains generated by LRMs.

To address this limitation, we propose CAT (\textbf{C}onfidence-\textbf{A}daptive \textbf{T}hinking), an adaptive reasoning framework driven by the model’s intrinsic confidence. Inspired by recent works on the quality estimation from the model's internal token distributions \cite{deepthink_with_conf,Geng_survey,Fadeeva_Fact-Checking}, our main idea is to leverage self-certainty \cite{self-certainty} as
the intrinsic fine-grained indicator to distinguish high-quality reasoning trajectories from erroneous ones.
\textbf{Firstly}, CAT employs self-certainty as the model's intrinsic confidence metric to 
estimate the quality of generated reasoning trajectories, which reflects the problem difficulty. 
Based on the separation of confidence and lengths between different trajectories, we further construct preference data to 
make the model capture the relationship between problem difficulties and output lengths.
\textbf{Secondly}, we devise a confidence-weighted preference optimization (CWPO) method, which weights the vanilla preference optimization objective with confidence.
This encourages the model to compress reasoning steps under high confidence
while retaining necessary exploration otherwise,
thereby mitigating overthinking for simple cases and maintaining reasoning performance especially for hard ones.

In summary, our main contributions are\footnote{Our codes are available at \url{https://github.com/Jiang9732/CAT-code}.}:
\begin{itemize}
  \item We introduce the confidence-adaptive thinking (CAT) framework that shifts the paradigm of efficient reasoning from external supervision to intrinsic confidence awareness. CAT enables reasoning models to autonomously perceive problem difficulty and modulate their thinking depth.
  \item We propose the confidence-weighted preference optimization (CWPO) objective 
  that dynamically weights the vanilla objective based on the calibration ratio of confidence to length. CWPO mitigates overthinking while preserving the model's ability to explore complex reasoning paths if necessary.
  \item We conduct extensive experiments across three challenging benchmarks and show superior performance of CAT over state-of-the-art baselines on the balance between inference efficiency and reasoning accuracy.
\end{itemize}

\section{Related Work}

\noindent\textbf{Efficient Reasoning in LRMs.} Recent studies have increasingly focused on the phenomenon of overthinking in large reasoning models \cite{Sui_survey,when_more_is_less,wang_relatedwork}. Existing efficient reasoning methods can generally be categorized into two streams. The first involves training strategies to equip LRMs with the ability to generate concise reasoning chains, spanning from supervised fine-tuning \cite{Cui_Perplexity-Guided,xia-etal-2025-tokenskip} to reinforcement learning \cite{DAST,Aggarwal_RL_L1,O1-Pruner,DAPO_RL}.
The second category comprises inference-time methods, including prompting \cite{han_prompt_method,Renze_and_Guven_prompt,nayab_prompt}, task routing \cite{conf_task_routing,routellm}, latent space compression \cite{Hao_latent,Codi}, and dynamic decoding \cite{sun_decoding,zhang2025confidence}. 

Compared with existing works on training methods of efficient reasoning, our work utilizes the model's confidence as the estimation of problem difficulty, instead of solely depending on
external reward models and extrinsic metrics. 
This makes the full usage of the model's intrinsic property to achieve adaptive reasoning.

\noindent\textbf{Confidence Utilization in LRMs}.
 Recent works have shown that the model confidence potentially indicate the quality of reasoning chains \cite{deepthink_with_conf,Geng_survey,self-certainty,Fadeeva_Fact-Checking}. As one of the representative metrics to reflect confidence, self-certainty \cite{self-certainty} has been primarily applied to Best-of-N selection \cite{deepthink_with_conf}.
 For comparison, our work uses self-certainty as the model's confidence to self-evaluate the quality of generated reasoning chains, which guides the adaptive thinking via preference optimization, instead of merely injecting it into the inference stage.

\begin{figure*}[t]
  \centering
  \includegraphics[width=\textwidth]{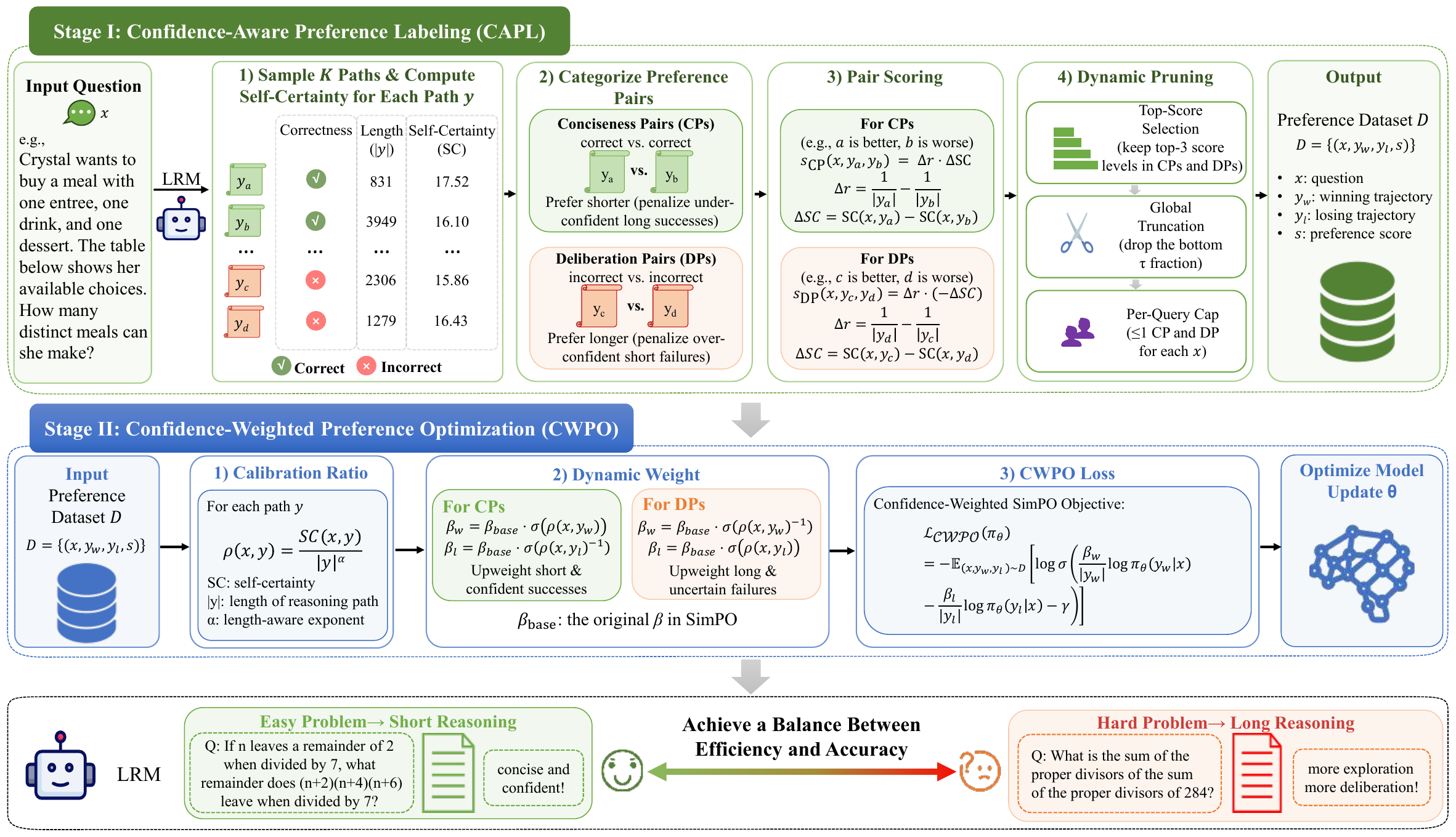}
  \caption{Overview of the CAT framework. 
  }
  \label{fig:main}  
\end{figure*}

\section{Methodology}
\subsection{Task Definition and Method Overview}

Given an input question $x$, our goal is to acquire a reasoning trajectory $y$ that contains a multi-step reasoning process and a final answer. Under the precondition of accuracy, $y$ is required to become short for simple problems while being long for hard ones if necessary.

An overview of our framework is presented in Figure~\ref{fig:main}. Firstly, we sample multiple reasoning trajectories for each question and compute their path-level self-certainty scores as confidence via a dedicated forward pass (Section \ref{sec:CAPL}). Secondly, we construct preference pairs based on the confidence and lengths,
and apply dynamic selection to prioritize more informative supervision (Section \ref{sec:CAPL}). 
Finally, we fine-tune the base LRM with a confidence-weighted preference optimization objective, which incorporates confidence and lengths to further modulate the preference strength, achieving conditional length regulation (Section \ref{subsubsec:cwpo}).

\subsection{Confidence-Adaptive Thinking}

Our confidence-adaptive thinking framework consists of two stages, including confidence-aware preference labeling and confidence-weighted preference optimization. While the first stage aims to incorporate confidence as intrinsic signals to construct fine-grained preference pairs, the second stage further utilizes confidence to further enhance the preference optimization objective.

\subsubsection{Confidence-Aware Preference Labeling}\label{sec:CAPL}

To build the preference dataset, we first sample $K$ reasoning trajectories $\{y^{(k)}\}_{k=1}^{K}$ for the question $x$ from the base reasoning model, each of which
is a token sequence $y^{(k)}=\left(y^{(k)}_1,\dots,y^{(k)}_{n_k}\right)$ with the length of $n_k$. 
The goal of this stage is to construct a preference dataset $\mathcal{D}=\{(x, y_w, y_l, s)\}$, where $y_w$ and $y_l$ denote the winning and losing trajectories for the same input $x$, and $s$ indicates the confidence-calibrated preference score.

\paragraph{Self-Certainty as Intrinsic Confidence.}
To capture the model's intrinsic confidence during reasoning, we follow \citet{self-certainty} to employ self-certainty, 
which can also serve as a trajectory-level quality measure.
Formally, assuming that $\bm{p}_\theta(\cdot \mid x, y_{\le i})$ denotes the next-token distribution at $i$-th position, $V$ indicates the vocabulary size, and $\mathcal{U}$ represents the uniform distribution over $V$, self-certainty (SC) can be computed as follows:
\begin{equation}
\label{eq:sc_definition}
\text{SC}(x,y)\;=\;-\frac{1}{nV}\sum_{i=1}^{n}\sum_{j=1}^{V}\log\!\Big(V\cdot \bm{p}_\theta(j\mid x,y_{\le i})\Big)
\end{equation}
which corresponds to measuring the KL divergence 
$D_{\mathrm{KL}}\!\bigl(\mathcal{U}\,\|\,p_\theta(\cdot\mid x, y_{\le i})\bigr)$ and averaging this quantity over $i$.
Intuitively, a larger divergence from the uniform distribution implies a more peaked (and thus more certain) predictive distribution, leading to higher SC. Conversely, a distribution closer to uniform is flatter, indicating greater uncertainty and yielding lower SC.

\paragraph{Preference Pair Construction.}
We consider three important factors of each trajectory to construct the preference dataset:
(i) the correctness of the answer,
(ii) the length, and (iii) the intrinsic confidence based on SC in Eq.~(\ref{eq:sc_definition}).
We emphasize that SC is a complementary to external factors, which estimates trajectory-level fine-grained qualities and determines the strength of pairwise preferences.

Inspired by \citet{DAST}, we categorize preference pairs into two types:
\textbf{Conciseness Pairs (CPs)}, formed by two correct trajectories where the preferred one is shorter;
and \textbf{Deliberation Pairs (DPs)}, formed by two incorrect traces where the preferred one is longer.
Unlike prior approaches that calibrate preference strength using per-question fixed budgets or external difficulty estimation, CAT uses only model-internal evidence to modulate pairwise preference score $s$.

For each input question $x$ and its $K$ candidate reasoning paths, we consider the margin between both lengths and self-certainty to acquire the preference score as $s$. Specifically,
given a candidate pair $(x, y_w, y_l)$, we first compute the margin in terms of self-certainty, lengths, and correctness:
\begin{equation}
\begin{aligned}
\Delta r & \;=\; r(y_w) - r(y_l) \\
\Delta \mathrm{SC} & \;=\; \mathrm{SC}(x,y_w) - \mathrm{SC}(x,y_l)
\end{aligned}
\end{equation}
where $\mathrm{SC}(\cdot)$ can be acquired by Eq.(\ref{eq:sc_definition}) and $r(\cdot)$ is a factor with respect to reasoning lengths and correctness:
\begin{equation}
\label{eq:path_score}
r(y) =
\begin{cases}
    +\frac{1}{|y|} & \text{if $y$ is correct} \\
    -\frac{1}{|y|} & \text{if $y$ is incorrect}
\end{cases}
\end{equation}
This design assigns the highest reward to short, correct paths while imposing the lightest penalty on long, incorrect paths. 
Conversely, short but incorrect paths receive the most severe penalty. 

For CPs, our intent is to favor short and confident solutions and reject long and unconfident ones.
We therefore multiply $r$ and $\mathrm{SC}$
so that a pair receives stronger strength precisely when the winning path $y_w$ is not only much more efficient but also more internally decisive:
\begin{equation}
s_{\text{CP}}(x,y_w,y_l) \;=\; \Delta r \cdot \Delta \mathrm{SC}
\label{eq:cp_score}
\end{equation}

For DPs, we want to prefer long and unconfident attempts over short and confident failures, discouraging premature yet decisive mistakes.
Accordingly, we reverse the confidence term, making wrong trajectories with larger certainty receive stronger penalties:
\begin{equation}
s_{\text{DP}}(x,y_w,y_l) \;=\; \Delta r \cdot \big(-\Delta \mathrm{SC}\big)
\label{eq:dp_score}
\end{equation}

In both cases, larger scores of preference pairs indicate potentially stronger and more discriminative preference signals for subsequent optimization. Thus, we devise a \textbf{Dynamic Pruning} strategy to select the preference optimization dataset based on $s$.
Concretely, for each query $x$, we rank the  CP and DP sets by their score $s$ in descending order, respectively, and retain only those pairs whose scores fall within the top three highest score levels.
We then pool candidates from all the queries and sort them globally based on the preference score $s$, truncating the list by removing the bottom $\tau$ fraction, where $\tau$ denotes the truncation ratio.
Finally, to prevent over-representing queries that produce many high-scoring pairs, we enforce a per-query cap and retain at most one CP and one DP per query in the final preference dataset.

\begin{table*}[t]
  \centering
  \fontsize{11}{12.5}\selectfont      
  \setlength{\tabcolsep}{3.2pt}       
  \renewcommand{\arraystretch}{1.25}
  \begin{adjustbox}{max width=\textwidth,center}
  \begin{tabular}{@{}lccccc ccccc ccccc@{}}
    \toprule
    \multirow{2}{*}{\textbf{Model}}
      & \multicolumn{5}{c}{\textbf{MATH 500}}
      & \multicolumn{5}{c}{\textbf{AIME 2024}}
      & \multicolumn{5}{c}{\textbf{GPQA}} \\
    \cmidrule(r){2-6}\cmidrule(lr){7-11}\cmidrule(l){12-16}
      & \textbf{Acc}$\uparrow$ & \textbf{Len}$\downarrow$ & \textbf{C-Len}$\downarrow$
      & \textbf{CR}$\uparrow$ & \textbf{C-CR}$\uparrow$
      & \textbf{Acc}$\uparrow$ & \textbf{Len}$\downarrow$ & \textbf{C-Len}$\downarrow$
      & \textbf{CR}$\uparrow$ & \textbf{C-CR}$\uparrow$
      & \textbf{Acc}$\uparrow$ & \textbf{Len}$\downarrow$ & \textbf{C-Len}$\downarrow$
      & \textbf{CR}$\uparrow$ & \textbf{C-CR}$\uparrow$ \\
    \midrule

    \rowcolor{gray!20}
    \textbf{R1-1.5B}
      & 82.9 & 5087  & 3595 & -- & --
      & 28.9 & 16609 & 8473 & -- & --
      & 34.0   & 10100 & 9240 & -- & -- \\
    OverThink$_{\text{SimPO}}$
      & 82.3 & 3899 & 2482   & 23.4\% & 31.0\%
      & 27.8   & 12104 & 6224 & 27.1\% & 26.6\%
      & 35.7 & 8908  & 7967 & 11.8\% & 13.8\% \\
    DAST
      & 85.5 & 4189 & 2635   & 17.7\% & 26.7\%
      & 30.0   & 13911 & 7667 & 16.2\% & 9.5\%
      & 36.7 & 9980  & 8993 & 1.2\% & 2.7\% \\
    ConCISE$_{\text{SimPO}}$
      & 80.8 & 3449  & 2300 & \textbf{32.2\%} & \textbf{36.0\%}
      & 25.6 & 10947 & 5663 & \textbf{34.1\%} & \textbf{33.2\%}
      & 34.9 & 7890  & 6995 & \textbf{21.9\%} & \textbf{24.3\%} \\
    \textbf{CAT (Ours)}
      & \textbf{86.1} & 4087  & 2736 & 19.7\% & 23.9\%
      & \textbf{31.1} & 12619 & 7155 & 24.0\% & 15.6\%
      & \textbf{37.0} & 9716  & 8348   & 3.8\% & 9.7\% \\
    \midrule

    \rowcolor{gray!20}
    \textbf{R1-7B}
      & 91.7 & 3657  & 3200 & -- & --
      & 53.3 & 12831 & 7631   & -- & --
      & 49.2 & 8553  & 7382 & -- & -- \\
    OverThink$_{\text{SimPO}}$
      & 89.6 & 2421 & 1873   & 33.8\% & 41.5\%
      & 52.2   & 10030 & 5845 & 21.8\% & 23.4\%
      & 49.8 & 7270  & 5856 & 15.0\% & 20.7\% \\
    DAST
      & 93.5 & 2997  & 2469 & 18.0\% & 22.8\%
      & 56.7 & 10804 & 7158 & 15.8\% & 6.2\%
      & 52.0   & 8068  & 7021   & 5.7\% & 4.9\% \\
    ConCISE$_{\text{SimPO}}$
      & 89.5 & 2187  & 1805 & \textbf{40.2\%} & \textbf{43.6\%}
      & 48.9 & 8854  & 5333 & \textbf{31.0\%} & \textbf{30.1\%}
      & 50.2 & 6030  & 5011 & \textbf{29.5\%} & \textbf{32.1\%} \\
    \textbf{CAT (Ours)}
      & \textbf{93.9} & 2431  & 1953 & 33.5\% & 39.0\%
      & \textbf{58.9} & 9880 & 5401 & 23.0\% & 29.2\%
      & \textbf{54.0} & 7781  & 6425   & 9.0\% & 13.0\% \\
    \midrule

    \rowcolor{gray!20}
    \textbf{Qwen3-8B}
      & 96.2   & 5549   & 5088 & -- & --
      & 74.4 & 14437   & 12060  & -- & --
      & 59.9 & 7241  & 6354 & -- & -- \\
    OverThink$_{\text{SimPO}}$
      & 94.9 & 3478 & 3107   & 37.3\% & 38.9\%
      & 73.3   & 10545 & 7960 & 27.0\% & \textbf{34.0\%}
      & 57.1 & 3712  & 3389 & \textbf{48.7\%} & \textbf{46.7\%} \\
    DAST
      & 96.5   & 4402    & 3964 & 20.7\% & 22.1\%
      & 75.6   & 13433 & 10252 & 7.0\% & 15.0\%
      & 59.9 & 6380  & 5523 & 11.9\% & 13.1\% \\   
    ConCISE$_{\text{SimPO}}$
      & 94.1 & 2880  & 2551   & \textbf{48.1\%} & \textbf{49.9\%}
      & 74.4 & 10190 & 7961 & \textbf{29.4\%} & \textbf{34.0\%}
      & 58.6 & 3939  & 3464 & 45.6\% & 45.5\% \\
    \textbf{CAT (Ours)}
      & \textbf{96.6} & 3546  & 3205 & 36.1\% & 37.0\%
      & \textbf{76.7} & 11774 & 9866 & 18.4\% & 18.2\%
      & \textbf{60.9} & 4819  & 4393   & 33.4\% & 30.9\% \\
    \bottomrule
  \end{tabular}
  \end{adjustbox}

  \caption{Accuracy (\textbf{Acc}), the mean response length over all trajectories (\textbf{Len}) and trajectories with correct final answers (\textbf{C-Len}), the percentage reduction in \textbf{Len} relative to the base model (\textbf{CR}), and the percentage reduction in \textbf{C-Len} relative to the base model (\textbf{C-CR}) on three benchmark datasets, respectively. }
  \label{tab:main}
\end{table*}

\subsubsection{Confidence-Weighted Preference Optimization (CWPO)} 
\label{subsubsec:cwpo}

To adjust the model's reasoning depth conditionally on its internal certainty, rather than applying a uniform length bias to all the samples,
we propose the \textbf{Confidence-Weighted Preference Optimization (CWPO)}
objective that pioneers the use of intrinsic self-certainty directly within the alignment loss landscape. Compared with the vanilla SimPO objective \cite{simpo},
we dynamically 
modulated the scaling factors of the winning and losing terms. Formally, the CWPO loss is computed as:

\begin{equation}
\label{eq:CWPO_loss}
\begin{aligned}
&\mathcal{L}_{\text{CWPO}}(\pi_\theta)
= -\mathbb{E}_{(x,y_w,y_l)\sim \mathcal{D}}
\Bigl[
\log \sigma\Bigl(\\
&\tfrac{\beta_w}{|y_w|}\log \pi_\theta(y_w|x)
-\tfrac{\beta_l}{|y_l|}\log \pi_\theta(y_l|x)
-\gamma
\Bigr)
\Bigr]
\end{aligned}
\end{equation}
where the dynamic weights $\beta_w$ and $\beta_l$ are acquired by the original $\beta_{base}$ in SimPO and a \textbf{calibration ratio} ($\rho$) based on self-certainty and lengths:
\begin{equation}
\rho(x,y) \;=\; \frac{\text{SC}(x,y)}{|y|^{\alpha}}
\label{eq:calibration_ratio}
\end{equation}
where $\alpha \in (0,1)$ is a length-aware exponent to keep $\mathrm{SC}$ and $|y|$ on a comparable scale for numerical stability. This ratio imposes an additional tunable length penalty so that confidence-guided scaling can better align gradient allocation with efficiency. 

The CWPO loss sets different weights for conciseness pairs (CPs) and deliberation pairs (DPs):
For CPs, where the model compares two correct responses, we define $\beta_w = \beta_{base} \cdot \sigma(\rho(x,y_w))$, while symmetrically scaling the loser's weight using the inverse ratio $\beta_l = \beta_{base} \cdot \sigma(\rho(x,y_l)^{-1})$. This specifically incentivizes the model to commit to reasoning paths that are both correct and concise. Conversely, for DPs, we focus on penalizing short and erroneous answers with unearned confidence. We set the penalty weight $\beta_l = \beta_{base} \cdot \sigma(\rho(x,y_l))$ and the winner's weight $\beta_w = \beta_{base} \cdot \sigma(\rho(x,y_w)^{-1})$. By integrating these internal signals, CWPO moves beyond static length penalties, allowing the model to autonomously judge when to compress reasoning and when to deliberate, 
achieving a balance between efficiency and accuracy.

\section{Experiments}

\subsection{Settings}
\noindent\textbf{Models and Datasets.} 
We conduct comparative experiments on three LRMs: DeepSeek-R1-Distill-Qwen-7B (R1-7B) / 1.5B (R1-1.5B) \cite{Deepseek-r1} and Qwen3-8B \cite{qwen3}. For the training dataset, following \citet{concise}, we randomly select 2,000 questions from the MATH training set \cite{math}, maintaining diversity in both difficulty and response length.

\noindent\textbf{Benchmarks.} 
We follow \citet{DAST} to select three benchmarks, including MATH-500 \cite{lets_verify_stepbystep}, AIME24 \cite{aime2024}, and GPQA \cite{gpqa}.

\noindent\textbf{Baselines.} 
We select several state-of-the-art methods
for efficient reasoning as baselines, including \textbf{OverThink} \cite{overthink}, \textbf{DAST} \cite{DAST}, and \textbf{ConCISE} \cite{concise}. For ConCISE, we choose the best-performing alternative ConCISE$_{\text{SimPO}}$ as the comparison baseline.
All of these methods are under the paradigm of preference optimization with the SimPO objective.

\noindent\textbf{Implementation Details.} 
Following \citet{DAST}, we generate 20 candidate responses per question in our training set, and set the maximum sequence length to 4,096 tokens. 
Based on the hyperparameter analysis in Appendix~\ref{app:analysis_hyper}, the truncation ratio ($\tau$) was set to 0.15.
The preference optimization is conducted
within the SimPO framework \cite{simpo}. We adopt low-rank adaption (LoRA) \cite{hu2022lora} with the rank of $r=32$, scaling factor of $\alpha=64$, dropout rate of 0.05. 
The training epoch is 1 while the batch size is 16.
The learning rate is 5e-5 for DeepSeek-R1-Distill-Qwen-7B and Qwen3-8B, and 5e-6 for DeepSeek-R1-Distill-Qwen-1.5B, as the weaker 1.5B backbone yields more DPs after CAPL and thus benefits from more conservative optimization. All the experiments are conducted on 
2 NVIDIA A800 GPUs. More training details are provided in Appendix~\ref{app:training_set}.
Decoding processes
are executed using the the OpenR1 evaluation scripts \cite{openr1},
with comprehensive decoding details provided in Appendix~\ref{app:eval_details}. The experimental results are presented with mean values over 3 runs.

\begin{figure*}[t]
  \begin{subfigure}{\columnwidth}
    \includegraphics[width=0.9\linewidth]{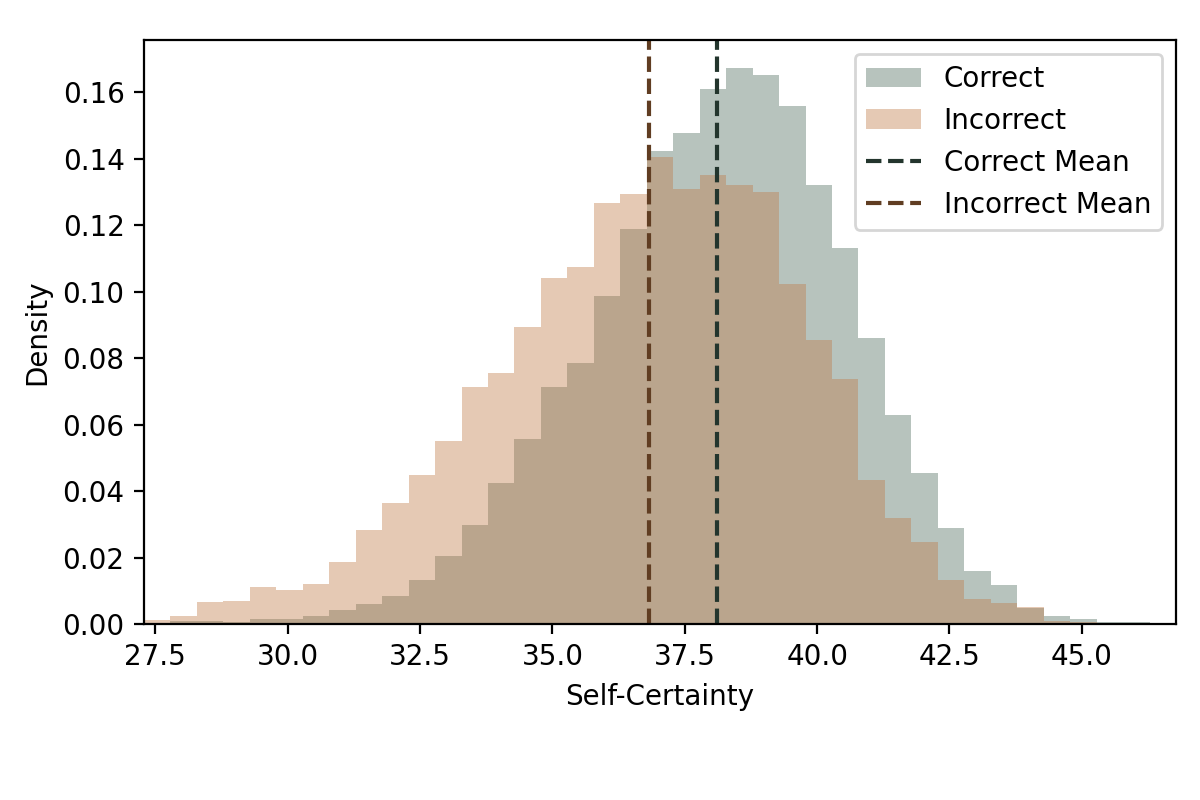}
    \caption{The distribution of SC for correct and incorrect responses.}
    \label{fig:sc_analysis_a}
  \end{subfigure}
  \hfill
  \begin{subfigure}{\columnwidth}
    \includegraphics[width=0.9\linewidth]{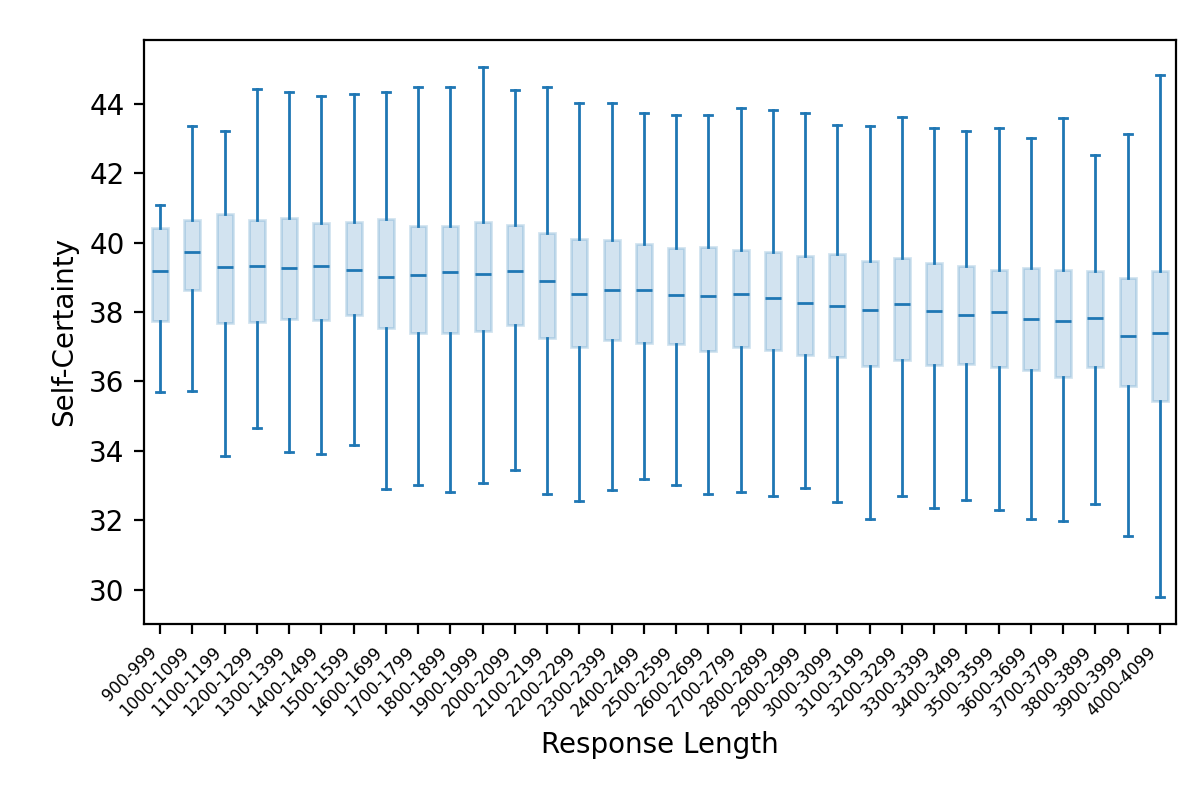}
    \caption{Box plots of SC across varying response lengths.}
    \label{fig:sc_analysis_b}
  \end{subfigure}
  \hfill
  \caption{Analysis of Self-Certainty (SC) distributions regarding response correctness and robustness to length on the MATH dataset (Level 4), derived from 20 reasoning paths per question generated by Qwen3-8B ($L_{max}$ = 4096). }
  \label{fig:sc_analysis}
\end{figure*}

\subsection{Results and Analysis}
\subsubsection{Overall Results}
The results in Table~\ref{tab:main} show that 
CAT achieves the highest accuracy (exceeding the backbone model) on all the three benchmarks while maintaining an acceptable compression rate (CR), suggesting that CAT can allocate reasoning steps adaptively to obtain better performance.
Although OverThink and ConCISE attain the most substantial compression rates, 
they still incur an unavoidable loss in accuracy relative to the backbone model. 
DAST and CAT exhibit similar balancing trends between task performance and compression, as both aim to achieve adaptive compression while preserving the model’s reasoning capability. Compared with DAST, CAT delivers higher for all the base models, and achieves higher CR and C-CR in most settings. These results suggest that CAT is more effective at adaptive reasoning, demonstrating effectiveness of our proposed confidence-aware adaptive reasoning approach based on the model’s intrinsic signals.

\begin{table}
\centering
\small
\setlength{\tabcolsep}{4pt}
\renewcommand{\arraystretch}{1.1}

\begin{adjustbox}{max width=\linewidth}
\begin{tabular}{c c c c c c c}
\toprule
Benchmark & Method & Acc & Len & C-Len & CR & C-CR \\
\midrule

\multirow{4}{*}{MATH-500}
& Origin            & 91.7 & 3657   & 3200 & -- & -- \\
& \textbf{CAT}  & \textbf{93.9} & 2431  & 1953 & \textbf{33.5\%} & \textbf{39.0\%} \\
& w/o CWPO         & 93.6 & 2869   & 2361  & 21.5\% & 26.2\% \\
& w/o CAPL & 93.0 & 2512 & 2014  & 31.3\% & 37.1\% \\
\midrule

\multirow{4}{*}{AIME24}
& Origin            & 53.3 & 12831 & 7631 & -- & -- \\
& \textbf{CAT}  & \textbf{58.9} & 9880 & 5401 & 23.0\% & \textbf{29.2\%} \\
& w/o CWPO         & 53.3 & 10859  & 6387 & 15.4\% & 16.3\% \\
& w/o CAPL & 56.7 & 9131 & 5807 & \textbf{28.8\%} & 23.9\% \\
\midrule

\multirow{4}{*}{GPQA}
& Origin            & 49.2 & 8553 & 7382 & -- & -- \\
& \textbf{CAT}  & \textbf{54.0} & 7781  & 6425   & 9.0\% & 13.0\% \\
& w/o CWPO         & 51.5 & 7877 & 6343 & 7.9\% & \textbf{14.1\%} \\
& w/o CAPL & 50.5 & 7745 & 6412 & \textbf{9.4\%} & 13.1\% \\
\bottomrule
\end{tabular}
\end{adjustbox}

\caption{Ablation study of Confidence-Aware Preference Labeling and Confidence-Weighted Preference Optimization on DeepSeek-R1-Distill-Qwen-7B.}
\label{tab:ablation_CAT}
\end{table}

\subsubsection{Ablation Study}
To assess the key components in CAT, including Confidence-Aware Preference Labeling (CAPL) and Confidence-Weighted Preference Optimization (CWPO), we conduct an detailed ablation study by removing either CAPL (w/o CAPL, where preferences are scored only by $\Delta r$) or CWPO (w/o CWPO, where we replace CWPO with vanilla SimPO). 
The results in Table~\ref{tab:ablation_CAT} show that all these parts contribute to the final performance.
We observe that constructing preference pairs solely based on length differences (w/o CAPL) yields a higher compression ratio but leads to a larger degradation in reasoning performance on most tasks. This observation highlights the importance of high-quality training data that provides difficulty-adaptive reasoning signals.

Due to the paper limit, we further explore the effect of Self-Certainty on CAT in Appendix \ref{app:ablationsc}.

\begin{figure*}[t]
  \centering
  \includegraphics[width=\textwidth]{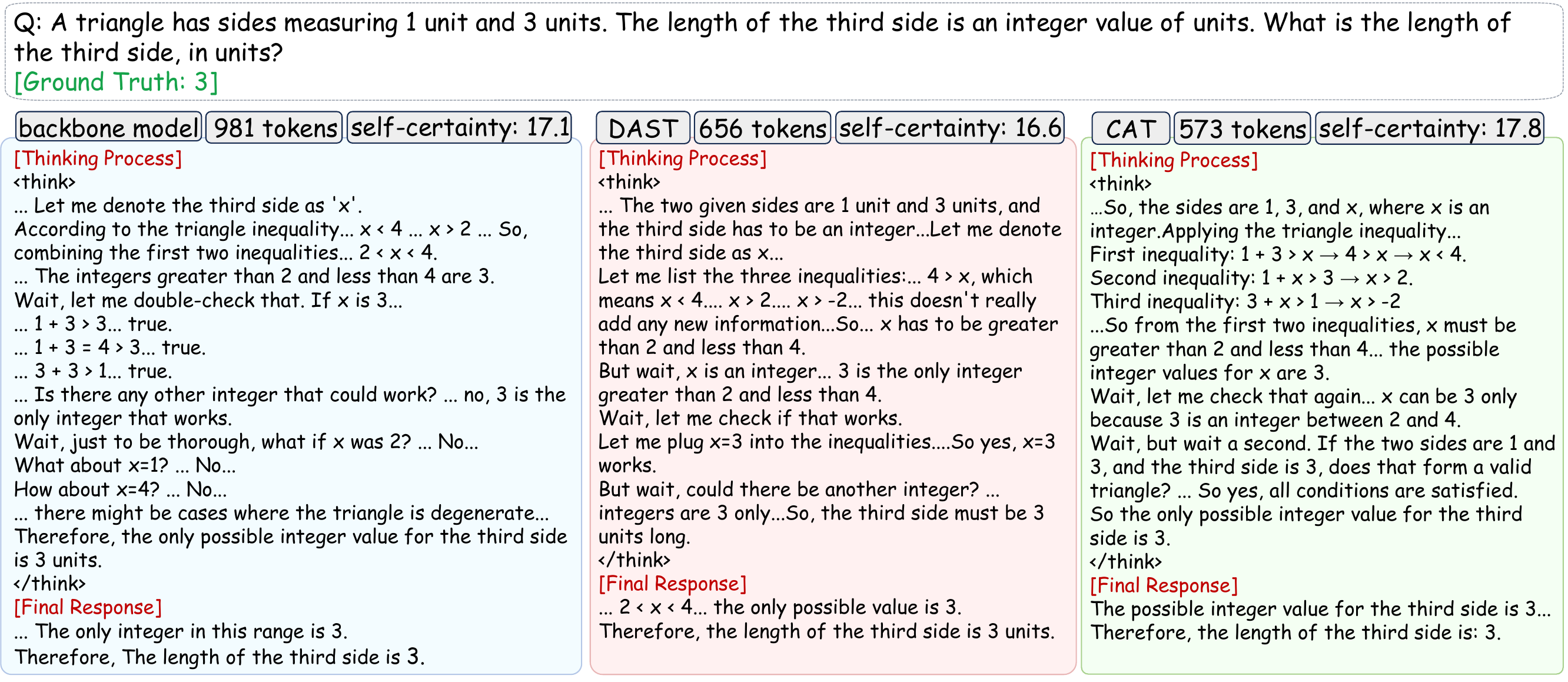}
  \caption{Case study on DeepSeek-R1-Distill-Qwen-7B. All the three methods solve the problem correctly. Compared with the backbone model and DAST that shorten the reasoning chain but lower self-certainty, 
  CAT further reduces the reasoning length and yields higher self-certainty.}
  \label{fig:case_study}  
\end{figure*}

\subsubsection{Analysis of Self-Certainty}
To better understand how self-certainty helps the model achieve an optimal balance between reasoning accuracy and length, we conduct a detailed analysis on the reasoning trajectories generated by Qwen3-8B 
of the MATH dataset.

\noindent\textbf{SC effectively distinguishes correct and incorrect reasoning paths.}
In Figure~\ref{fig:sc_analysis_a}, we analyze the distribution of self-certainty for correct and incorrect responses on MATH (Level 4). The distributions for correct and incorrect responses concentrate around distinct means, with correct responses exhibiting a higher mean.
This suggests that SC can effectively distinguish correct from incorrect reasoning trajectories and is strongly correlated with response quality.

\noindent\textbf{SC is robust to reasoning lengths.}
We analyze self-certainty across responses of varying lengths. As illustrated in Figure~\ref{fig:sc_analysis_b}, SC is not noticeably affected by response lengths: across the entire length range, the median (blue line) shows only a very slight downward trend, which is largely attributable to the fact that shorter responses contain a higher proportion of correct trajectories, 
indicating that SC is stable regardless of reasoning lengths.

\subsubsection{Generalization Across Preference Optimization Methods}

To test the generalization ability of our method,
we further apply our method to DPO in addition to SimPO,
and assess the performance on DeepSeek-R1-Distill-Qwen-7B using the same preference pairs constructed by CAPL. The DPO-version CWPO objective (denoted as CWPO$_{\text{DPO}}$) is slightly different from vanilla CWPO, which is detailed in Appendix \ref{app:cwdpo}.
The results in Table~\ref{tab:CWDPO} indicate that CWPO$_{\text{DPO}}$ beats standard DPO in most of the metrics on three benchmarks, demonstrating the promising generalization ability to different preference optimization methods.

\begin{table}
\centering
\small
\setlength{\tabcolsep}{4pt}
\renewcommand{\arraystretch}{1.1}

\begin{adjustbox}{max width=\linewidth}
\begin{tabular}{c c c c c c c}
\toprule
Benchmark & Method & Acc & Len & C-Len & CR & C-CR \\
\midrule

\multirow{3}{*}{MATH-500}
& Origin            & 91.7 & 3657   & 3200 & -- & -- \\
& DPO  & 92.7 & 2827 & 2188  & 22.7\% & 31.6\% \\
& \textbf{CWPO$_{\text{DPO}}$}        & \textbf{93.8} & 2722   & 2134  & \textbf{25.6\%} & \textbf{33.3\%} \\
\midrule

\multirow{3}{*}{AIME24}
& Origin            & 53.3 & 12831 & 7631 & -- & -- \\
& DPO  & 53.3 &10187  & 6145  & \textbf{20.6\%} & 19.5\% \\
& \textbf{CWPO$_{\text{DPO}}$}         & \textbf{58.9} & 10720 & 5532  & 16.5\% & \textbf{27.5\%} \\
\midrule

\multirow{3}{*}{GPQA}
& Origin            & 49.2 & 8553 & 7382 & -- & -- \\
& DPO & 50.7 & 8353 & 7057  & 2.3\% & 4.4\% \\
& \textbf{CWPO$_{\text{DPO}}$}        & \textbf{53.0} & 8326   & 6864  & \textbf{2.7\%} & \textbf{7.0\%} \\
\bottomrule
\end{tabular}
\end{adjustbox}

\caption{Results of DPO and CWPO$_{\text{DPO}}$ on DeepSeek-R1-Distill-Qwen-7B.}
\label{tab:CWDPO}
\end{table}

\subsubsection{Case Study}

To intuitively illustrate how CAT affects reasoning behaviors, 
we present a case study on DeepSeek-R1-Distill-Qwen-7B in Figure~\ref{fig:case_study}.
We observe that all the three methods reach the correct answer but exhibit different reasoning lengths and self-certainty.
The backbone model repeatedly verifies the same inequalities and explicitly checks invalid alternatives. 
DAST reduces the reasoning length with lower self-certainty, while still retaining additional verification beyond the core derivation.
In comparison, CAT achieves the shortest reasoning chain with higher self-certainty. It does not eliminate reflection entirely: after deriving the feasible interval, it only keeps a brief validity check 
rather than exploring invalid candidates. This qualitative case supports the design of CAT, which incorporates self-certainty together with correctness and length signals to favor concise and confident correct reasoning paths.

\section{Conclusion}
This work proposes confidence-adaptive thinking (CAT), 
which addresses the pronounced overthinking and token overhead in large reasoning models through intrinsic confidence awareness. CAT integrates self-certainty as LRMs' intrinsic confidence to enable them to 
compress confident responses while deliberating on uncertain ones. Extensive experiments demonstrate that CAT consistently achieves a superior balance between inference efficiency and accuracy.

\section*{Limitations}
While CAT demonstrates a superior balance between reasoning accuracy and efficiency, we identify the following areas for future improvement:

\paragraph{Path-Level Aggregation.}
Our current framework utilizes path-level Self-Certainty to score reasoning traces. While this metric effectively differentiates high-quality responses, aggregating token-level signals into a single scalar for the entire sequence may overlook variations in confidence at specific reasoning steps. Future work could explore integrating token positions with their specific Self-Certainty scores to enable more precise step-level compression.

\paragraph{Domain-Specific Evaluation.}
Our experiments focus on STEM disciplines such as mathematics and physics that allow rigorous correctness verification. Although Self-Certainty is an intrinsic signal independent of ground truth, our preference labeling strategy currently utilizes verification results. We aim to extend this approach to open-ended generation tasks where Self-Certainty can guide alignment without reliance on external answers.

\paragraph{Offline Optimization Paradigm.}
CAT employs Confidence-Weighted Preference Optimization on static datasets constructed from pre-sampled trajectories. This offline setting limits the ability of the policy to dynamically update its confidence estimates during the training process. Future research will investigate transitioning from offline optimization to online reinforcement learning variants, allowing the model to iteratively refine its reasoning efficiency through continuous interaction.

\section*{Acknowledgments}
This work was supported by Sichuan Science and Technology Program (2025ZNSFSC1488), Noncommunicable Chronic Diseases-National Science and Technology Major Project (2023ZD0501806), Fundamental Research Funds for the Central Universities (ZYGX2025XJ041), and CIPS-SMP-Zhipu Large Model Fund (CIPS-SMP20250314).

\bibliography{references}

\appendix

\section{Experimental Details}

\begin{figure*}[t]
  \begin{subfigure}{\columnwidth}
    \includegraphics[width=\linewidth]{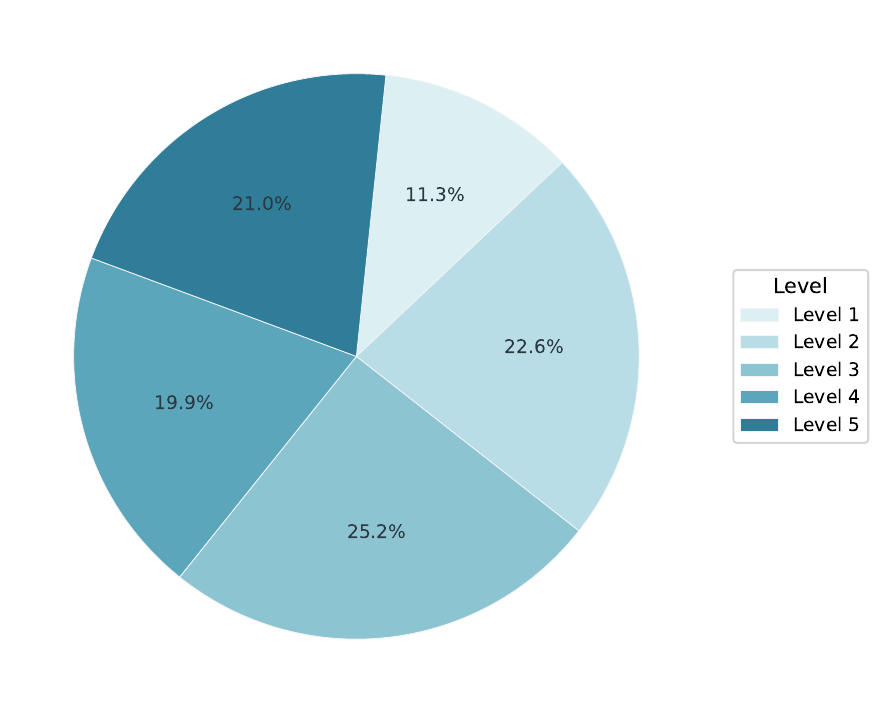}
    \caption{Difficulty diversity of training set.}
    \label{fig:dataset_level}
  \end{subfigure}
  \hfill
  \begin{subfigure}{\columnwidth}
    \includegraphics[width=\linewidth]{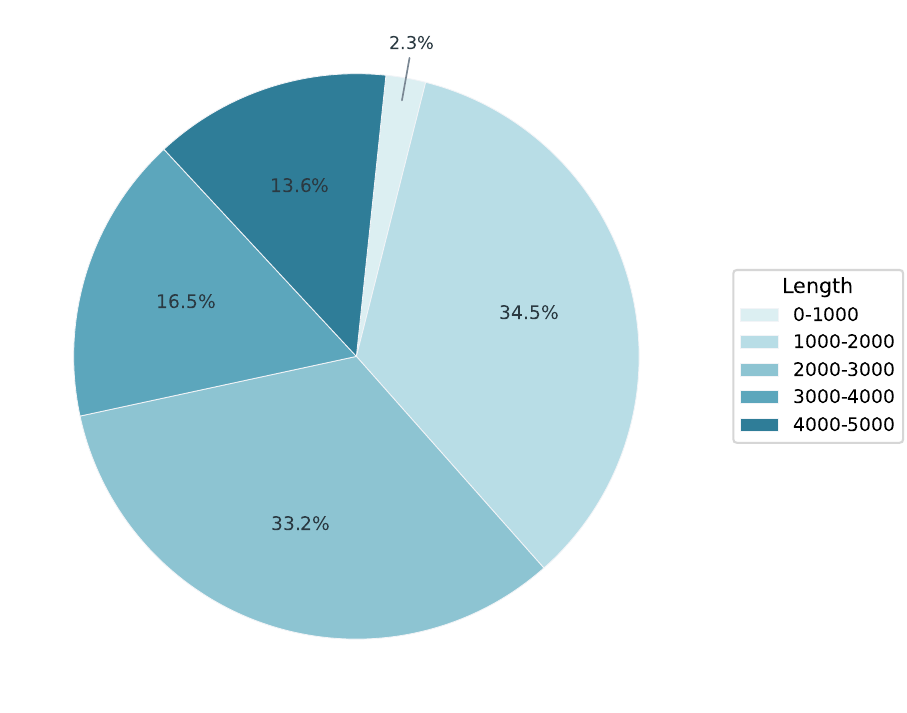}
    \caption{Length diversity of training set.}
    \label{fig:dataset_len}
  \end{subfigure}
  \hfill
  \caption{Difficulty and length distributions illustrating the diversity of the question set.}
  \label{fig:dataset}
\end{figure*}

\subsection{Training details}
\label{app:training_set}
\paragraph{Training Set.}
While the construction methodology of the training set is detailed in Section 4, Figures~\ref{fig:dataset_level} and ~\ref{fig:dataset_len} illustrate its diversity in terms of difficulty and length. Specifically, the difficulty distribution aligns with the \textit{Level} metric of the 2000 selected questions from the MATH dataset. The length distribution reflects the reasoning chains generated by DeepSeek-R1-Distill-Qwen-7B, obtained by sampling 20 paths per question with a temperature of 0.6 and a top\_p of 0.95.
\paragraph{Training Configurations.}
Following Dynamic Pruning with a truncation ratio of $\tau=0.15$, our constructed preference pairs for R1-7B yielded 1,765 Conciseness Pairs (CPs; 93.1\%) and 130 Deliberation Pairs (DPs; 6.9\%), a distribution similar to that reported in DAST \cite{DAST}. Similarly, for R1-1.5B, we identified 1,742 CPs (88.8\%) and 221 DPs (11.2\%). For Qwen3-8B, the resulting dataset comprised 1,517 CPs (97.4\%) and 40 DPs (2.6\%).
These results indicate that for models with stronger reasoning capabilities, such as Qwen3-8B, Confidence-Aware Preference Labeling generates a higher proportion of CPs to facilitate the learning of conciseness. Conversely, for models with weaker reasoning abilities, such as R1-1.5B, the method produces more DPs to encourage cautious exploration. Furthermore, following DAST, the original SimPO hyperparameters were set to $\beta=200$ and $\gamma=1$ for all three models.
All baselines use comparable data budgets to CAT.

\subsection{Evaluation Details} 
\label{app:eval_details}
In our evaluation setup, we use a unified decoding configuration for all experiments, with temperature = 0.6 and top\_p = 0.95 \cite{Deepseek-r1}. The maximum generation length is capped at 32,768 tokens for all the models.

\begin{figure*}
  \begin{subfigure}{\columnwidth}
    \includegraphics[width=\linewidth]{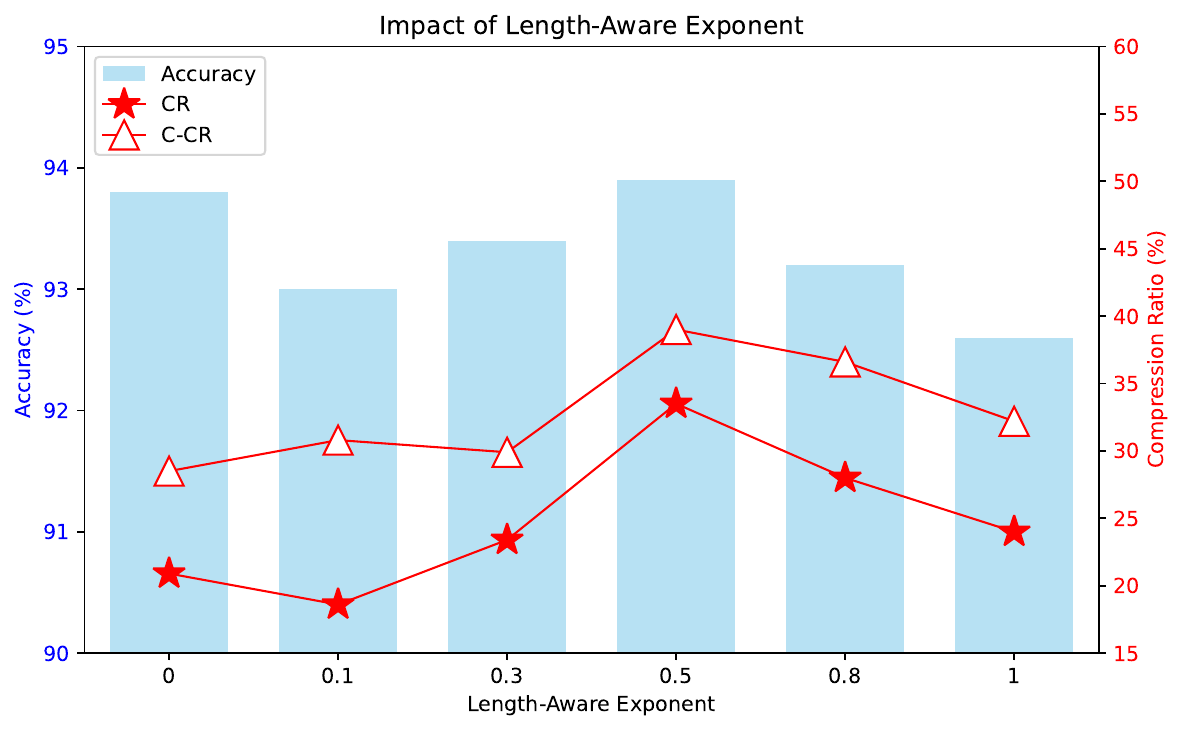}
    \caption{Impact of varying the Length-Aware Exponent $\alpha$ on model performance.}
    \label{fig:alpha}
  \end{subfigure}
  \hfill
  \begin{subfigure}{\columnwidth}
    \includegraphics[width=\linewidth]{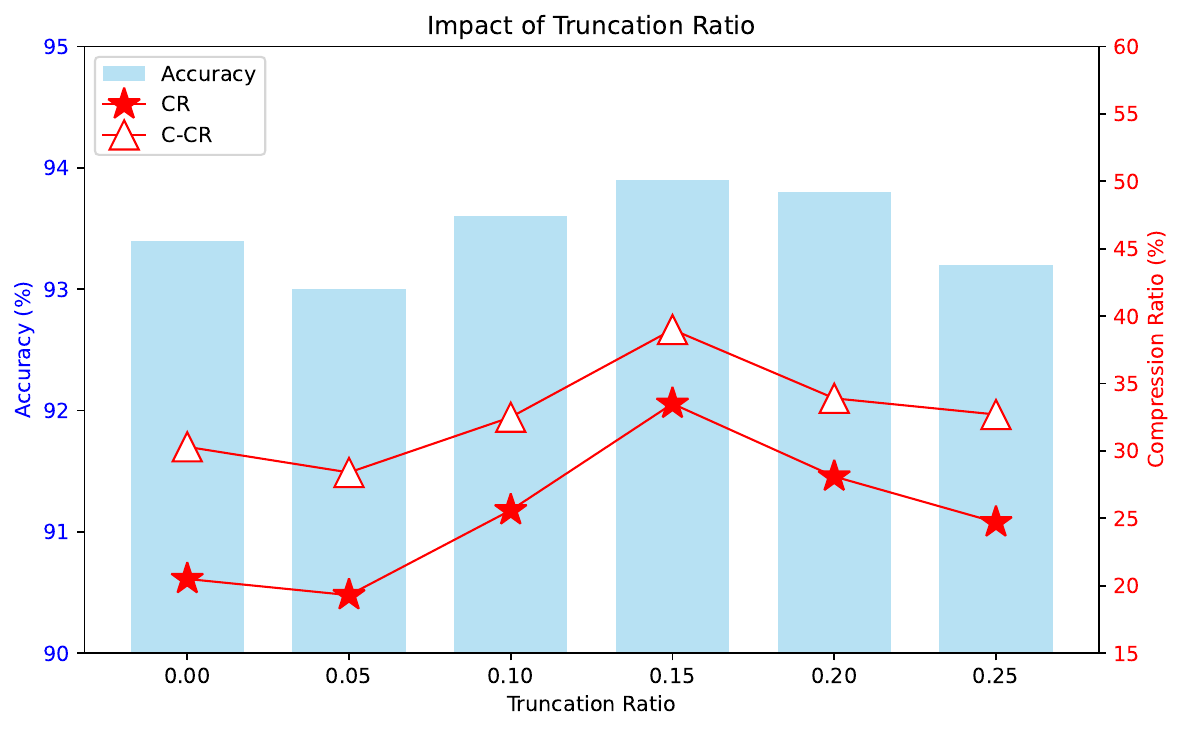}
    \caption{Impact of varying the Truncation Ratio $\tau$ on model performance.}
    \label{fig:tau}
  \end{subfigure}
  \hfill
  \caption{Hyperparameter analysis.}
  \label{fig:alpha_tau}
\end{figure*}

\begin{figure*}[t]
  \begin{subfigure}{\columnwidth}
    \includegraphics[width=\linewidth]{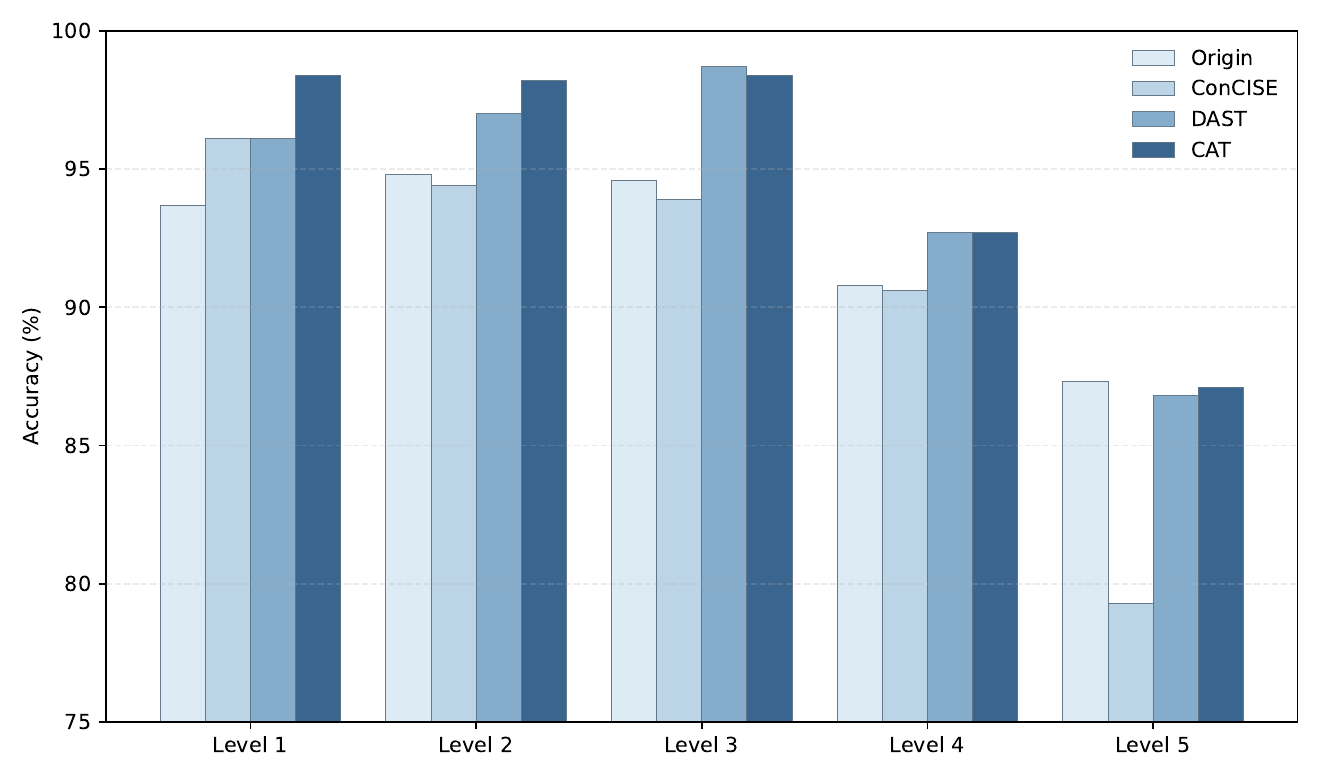}
    \caption{Accuracy Performance on Difficulty Levels.}
    \label{fig:baseline_level_acc}
  \end{subfigure}
  \hfill
  \begin{subfigure}{\columnwidth}
    \includegraphics[width=\linewidth]{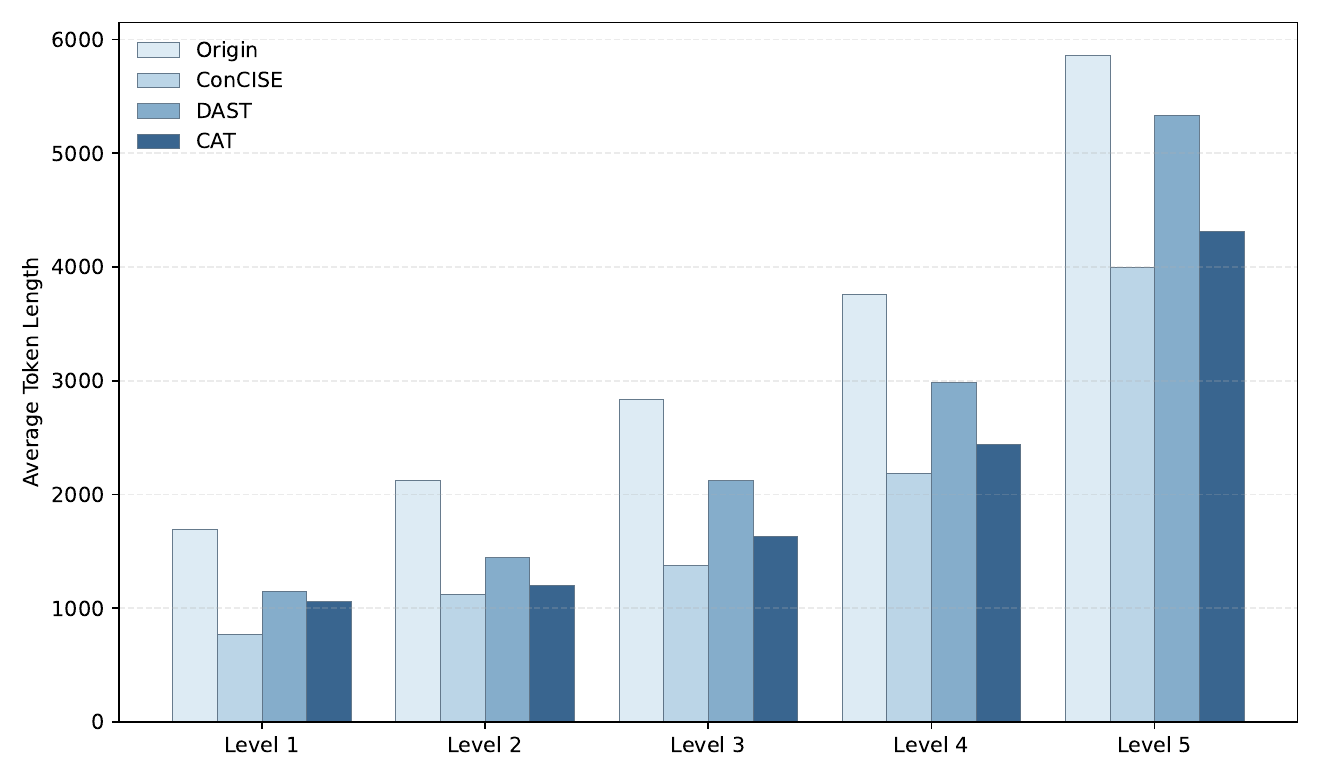}
    \caption{Average Token Length on Difficulty Levels.}
    \label{fig:baseline_level_len}
  \end{subfigure}
  \hfill
  \caption{Performance comparison on MATH-500 across different difficulty levels.}
  \label{fig:baseline_level}
\end{figure*}

\section{Additional Experiments }

\subsection{Ablation Study on Self-Certainty}
\label{app:ablationsc}

\begin{table}
\centering
\small
\setlength{\tabcolsep}{4pt}
\renewcommand{\arraystretch}{1.1}

\begin{adjustbox}{max width=\linewidth}
\begin{tabular}{c c c c c c c}
\toprule
Benchmark & Method & Acc & Len & C-Len & CR & C-CR \\
\midrule

\multirow{4}{*}{MATH-500}
& Origin            & 91.7 & 3657   & 3200 & -- & -- \\
& \textbf{CAT}  & \textbf{93.9} & 2431  & 1953 & \textbf{33.5\%} & 39.0\% \\
& w/o SC in CWPO only & 93.2 & 2771 & 2218  & 24.2\% & 30.7\% \\
& w/o SC in CAPL \& CWPO	 & 92.6 & 2511 & 1859  & 31.3\% & \textbf{41.9\%} \\
\midrule

\multirow{4}{*}{AIME24}
& Origin            & 53.3 & 12831 & 7631 & -- & -- \\
& \textbf{CAT}  & \textbf{58.9} & 9880 & 5401 & 23.0\% & 29.2\% \\
& w/o SC in CWPO only & 53.3 & 9088 & 6237  & \textbf{29.2\%} & 18.3\% \\
& w/o SC in CAPL \& CWPO	 & 50.0 & 10507 & 4906  & 18.1\% & \textbf{35.7\%} \\
\midrule

\multirow{4}{*}{GPQA}
& Origin            & 49.2 & 8553 & 7382 & -- & -- \\
& \textbf{CAT}  & \textbf{54.0} & 7781  & 6425   & 9.0\% & 13.0\% \\
& w/o SC in CWPO only & 51.0 & 8036 & 6489  & 6.0\% & 12.1\% \\
& w/o SC in CAPL \& CWPO	 & 49.5 & 7561 & 6185  & \textbf{11.6\%} & \textbf{16.2\%} \\
\bottomrule
\end{tabular}
\end{adjustbox}

\caption{Ablation study of Self-Certainty (SC) on DeepSeek-R1-Distill-Qwen-7B.}
\label{tab:ablation_SC}
\end{table}

To further investigate the effects of Self-Certainty (SC) in CAT, we conduct an additional ablation study to remove SC in CWPO or CAPL \& CWPO, respectively. Note that removing SC in CAPL only is equivalent to w/o CAPL in Table \ref{tab:ablation_CAT}, which is not repeatedly analyzed in this section.

Table \ref{tab:ablation_SC} shows that SC matters in both data construction and model optimization.
Although removing SC commonly brings a higher compression rate, the reasoning accuracy largely degrades on all the benchmarks.
This directly supports that SC contributes non-redundant value to our method.

\begin{figure}[!t]
  \includegraphics[width=\columnwidth]{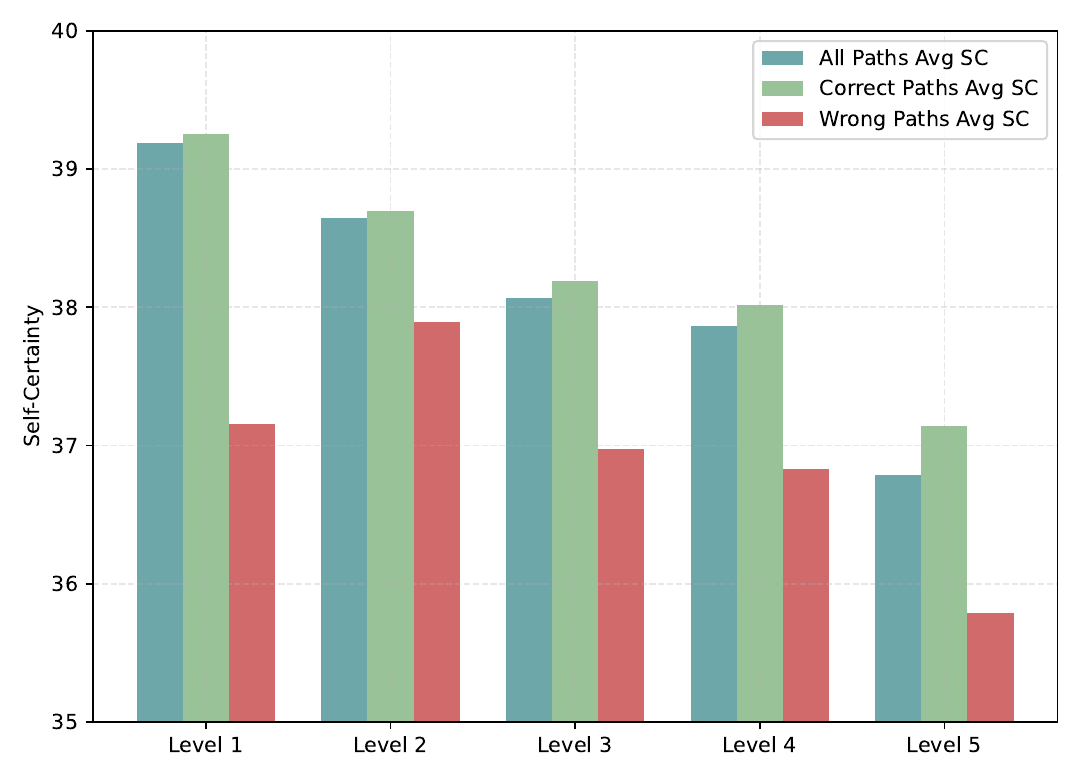}
  \caption {Average Self-Certainty (SC) across difficulty levels on the training set, derived from 20 reasoning paths per question generated by Qwen3-8B. A clear downward trend in SC is evident for all, correct, and incorrect trajectories as the difficulty level increases from Level 1 to Level 5. This indicates that SC effectively captures problem difficulty beyond correctness, serving as a reliable signal for difficulty-aware control.}
  \label{fig:sc_level}
\end{figure}

\subsection{Analysis of Hyperparameters}
\label{app:analysis_hyper}
To examine the effects of the Length-Aware Exponent($\alpha$) and the Truncation Ratio ($\tau$), we performed a grid search on MATH-500 using the R1-7B model for validation. As shown in Figure~\ref{fig:alpha_tau}, R1-7B achieves the optimal results when $\alpha$=0.5 and $\tau$=0.15. In both single-parameter sweeps, these settings yield the best Acc and compression ratio.
For the other backbone models, we conducted the same hyperparameter search and selected the final hyperparameters accordingly.

\subsection{Validation of SC as a Difficulty Indicator}
\label{app:sc_level}
Because MATH Level 4 does not provide a sufficiently diverse difficulty distribution, we analyze SC across different difficulty levels on the training set. As shown in Figure~\ref{fig:sc_level}, as the difficulty level increases, the average SC over all trajectories (including both correct and incorrect ones) exhibits a clear downward trend. This suggests that, beyond separating correct from incorrect trajectories, SC also captures differences in problem difficulty. This property is central to CAT: it enables difficulty-aware control.

\subsection{Performance Analysis Across Difficulty Levels}
We analyze the performance of CAT, DAST, and CONCISE on the MATH-500 dataset using the R1-7B model across varying difficulty levels. As shown in Figure~\ref{fig:baseline_level}. CAT achieves the highest accuracy in the two most challenging difficulty tiers, demonstrating a substantial advantage over other baselines and highlighting its robust reasoning capabilities for complex problems. While DAST also exhibits difficulty adaptability, it underperforms CAT in both accuracy and length compression. Furthermore, although CONCISE achieves the most significant length reduction, its reasoning performance deteriorates sharply as problem difficulty increases, creating a marked gap compared to the other methods and indicating a lack of capability in handling complex reasoning tasks.

\section{Details of CWPO$_{\text{DPO}}$ Objective}
\label{app:cwdpo}

To examine whether the benefits of CWPO-style SC weighting can generalize beyond the original CWPO setting, we incorporate the same SC-based dynamic weighting strategy into the DPO objective. 
The resulting objective, denoted as CWPO$_{\text{DPO}}$, is formulated as follows:
\begin{equation}
\label{eq:CWDPO_loss}
\begin{aligned}
&\mathcal{L}_{\text{CWPO$_{\text{DPO}}$}}(\pi_\theta; \pi_{\text{ref}})
= -\mathbb{E}_{(x,y_w,y_l)\sim \mathcal{D}}
\Bigl[
\log \sigma\Bigl(\\
&\beta_w \log \frac{\pi_\theta(y_w|x)}{\pi_{\text{ref}}(y_w|x)}
-\beta_l \log \frac{\pi_\theta(y_l|x)}{\pi_{\text{ref}}(y_l|x)}
\Bigr)
\Bigr]
\end{aligned}
\end{equation}
Here, we use the same definitions of $\beta_w$ and $\beta_l$ as in Section~\ref{subsubsec:cwpo}.

\end{document}